# ROC CURVES WITHIN THE FRAMEWORK OF NEURAL NETWORK ASSEMBLY MEMORY MODEL: SOME ANALYTIC RESULTS

Petro M. Gopych

***Abstract***: On the basis of convolutional (Hamming) version of recent Neural Network Assembly Memory Model (NNAMM) for intact two-layer autoassociative Hopfield network optimal receiver operating characteristics (ROCs) have been derived analytically. A method of taking into account explicitly a priori probabilities of alternative hypotheses on the structure of information initiated memory trace retrieval and modified ROCs (mROCs, a posteriori probabilities of correct recall vs. false alarm probability) are introduced. The comparison of empirical and calculated ROCs (or mROCs) demonstrates that they coincide quantitatively and in this way intensities of cues used in appropriate experiments may be estimated. It has been found that basic ROC properties which are one of experimental findings underpinning dual-process models of recognition memory can be explained within our one-factor NNAMM.

## 1. Introduction

Receiver operating characteristics (ROCs or ROC curves) are widely used in classic signal detection theory to provide the performance of linear Fisher or Euclidian classifiers for different values of their thresholds; ROCs plot the probability of correct detection of a noisy signal as a function of the probability of its false detection or false alarm [1]. Usually, it is assumed that distributions of initial patterns (vectors) conditioned on the presence or absence of the sought-after signal (prior probabilities of the both hypotheses are chosen to be ½) are Gaussians with the same (or similar) variances and a specific distance between them. In neurosciences, ROCs are used, for example, in data analysis where in single or multiple neuronal spike trains the encoding and processing of sensory information are studied, e.g., [2]. Lately, a method for deriving ROCs by means of human memory testing has been developed but up to present there exists no computer memory model which was able to reproduce empirical ROCs neither qualitatively nor quantitatively [3]. For this reason in the field of computer memory modeling understanding observed ROC curves is recognized as one of the most important unsolved problems [4].

In contrast to abstract computer models, neurobiology models directly address the problem of functional nature and neuroanatomical substrates of different kinds of memory. For example, now recognition memory is hotly debated within dual-process models (DPMs) which consider recognition as consisting of two components, recollection and familiarity, e.g., [3,5,6]. Recollection is thought of as an event where a person recalls both particular stimulus (a human face, for example) and episode where it was encountered earlier and familiarity represents the person's experience (or feeling) that particular stimulus was encountered before but without specific memory about where, when, or why it happened. It is claimed [3] that DPMs are supported by many results of cognitive, neuropsychological, and neuroimaging memory studies but in spite of long history of research even basic properties of DPMs are ambiguously defined and rather often even their basic terms are used by different authors in different ways [3]. Additionally, DPMs are not specified on computational level because most computer models consider recognition as one- not as two-factor process (although see [6]). On the other hand, none of computer models describes the whole body of recognition memory traits (in particular, ROCs) and for this reason their separate inferences which are not consistent with predictions of DPMs cannot be viewed as convincing arguments against them.

In present work analytical formulae for optimal ROC calculations are derived and, using convolutional (Hamming) version of Neural Network Assembly Memory Model (NNAMM) [7,8], we show that ROCs, as one of experimental findings underpinning DPMs, can be explained within our NNAMM without assuming that recognition memory is a dual process. A method of taking into account explicitly prior probabilities of alternative hypotheses on the structure of information initiating memory retrieval is proposed; on this basis modified ROC (mROC, unconditional probability of correct recall vs. false alarm probability) and overall probabilities of memory trace recall/recognition were introduced. It has been found that comparison of calculated and empirical ROCs (or mROCs) provides a method for extraction of those cues which were used actually in appropriate memory experiments.

## 2. Some NNAMM Backgrounds

According to NNAMM (see ref. 8 for details), components of initial ternary vectors take their values from the triple set –1,0,1 but most of these values are 0s (that is so called sparse coding). After data preprocessing, initial ternary vectors are transformed into binary feature vectors with components –1 or 1 (that is so called dense coding). In fact, feature vectors are quasibinary ones because their spinlike (–1,1) components cannot be shifted to other (0,1) binary representation and they could manifest (although do not manifest) their third, zero, components. Below only quasibinary vectors are considered but, for short, the preposition "quasi" will be omitted.

Neural network (NN) assembly memory is constructed from interconnected (associated) and equal in rights assembly memory units (AMUs) and the basic properties of assembly memory as a whole depend on the properties of its components, AMUs. AMU has original architecture and involves regular Hopfield two-layer autoassociative NN (that is the AMU's central element), $N$-channel time-gate, additional reference memory, and two nested feedback loops [8].

NN related to particular AMU is subserved by binary vectors mentioned. We refer to such an $N$-dimensional arbitrary vector as $x$. If $x$ represents information stored or that should be stored in AMU then we term it $x_0$. We define random vector or binary noise $x_r$ as $x$ with components –1 or 1 randomly chosen with uniform probability, ½. Damaged reference vector, $x(d)$, is defined as $x_0$ with its damage degree $d$. The components, $x_i(d)$, of $x(d)$ are defined as

$$x_i(d) = \begin{cases} x_0^i, & \text{if } u_i = 0, \\ x_r^i, & \text{if } u_i = 1 \end{cases} \quad i = 1,..,N \tag{1}$$

where $u_i$ are marks whose magnitudes 0 or 1 are chosen randomly with uniform probability and fixed d:

$$d = \sum u_i / N, \quad i = 1,..,N. \tag{2}$$

If the number of marks $u_i$ = 1 is $m$ then $d = m/N$; $0 \le d \le 1$; $x(0) = x_0$ and $x(1) = x_r$. Damage degree $d$ is a fraction of noise in vector $x(d)$ while intensity of cue or cue index $q = 1 - d$ is a fraction of correct, undamaged information about $x_0$ in $x(d)$ [7,8]. The data coded in such a way naturally arise when to solve a very important problem of local feature discrimination across smooth background and additive noise, line or half-tone images are binarized using a convolutional NN recognition algorithm [9]. Expressions 1 and 2 define an original data coding procedure [7]. To design appropriate data decoding rules we explore two-layer auto-associative NN with $N$ cells in its entrance (or exit) layer. Entrance and exit layer cells are connected by "all-to-all" rule, they are McCalloch-Pitts model neurons with rectangular response and triggering threshold $\theta$.

Following ref. 10 for perfectly learned intact Hopfield NN, the elements $w_{ij}$ of *synapse matrix w* are defined as

$$w_{ij} = \eta x_0^i x_0^j \tag{3}$$

where $i,j = 1,..,N$; $\eta > 0$ is a *learning parameter* (below $\eta = 1$); $x_0^i, x_0^j$ are the components of reference vector $x_0$ (all $w_{ij}$ may differ from each other in sign only). It is crucially important to stress that NN with synapse matrix $w$ is learned to remember *only one* memory trace $x_0$ and we *deliberately* reject the available possibility of storing other traces in the same NN. Also we posit that an input vector $x_{in}$ is decoded (recognized as reference vector $x_0$) successfully if learned NN transforms $x_{in}$ into output vector $x_{out} = x_0$ [7,8,9].

The transformation algorithm is the following. For the $j$th neuron of the NN exit layer an *input signal* $h_j$ is given by

$$h_j = \sum w_{ij} v_i + s_j \tag{4}$$

where $v_i$ is an *output signal* of the $i$th neuron of the NN entrance layer; $s_j = 0$.

The signal $v_j$ of the $j$th neuron of the NN exit layer (the $j$th component of $x_{out}$) is calculated according to the model neuron's rectangular response function (signum function or 1 bit quantifier) with triggering threshold $\theta$ as

$$v_j = \begin{cases} +1, & \text{if } h_j > \theta \\ -1, & \text{if } h_j \leq \theta \end{cases} \tag{5}$$

where for $h_j = \theta$ the value $v_j = -1$ was *arbitrary* assigned.

## 3. Convolutional and Hamming Versions of NNAMM

If $h_i = x_{in}^i$ then from Expression 5 follows that $v_i = x_{in}^i$. Of this fact and Equations 3 and 4 for the $j$th exit layer neuron we have: $h_j = \sum w_{ij} x_{in}^i = \eta x_0^j \sum x_0^i x_{in}^i = \eta x_0^j Q$ where $Q = \sum x_0^i x_{in}^i$ is a convolution of vectors $x_0$ and $x_{in}$ ($-N \leq Q \leq N$). The substitution of $h_j = \eta x_0^j Q$ into Expression 5 gives that $x_{out} = x_0$ and vector $x_{in}$ is successfully decoded (recognized as $x_0$) if $Q > \theta$ (if $\eta \neq 1$ then $Q > \theta/\eta$). Hence, NN algorithm given in Section 2 and the convolutional algorithm just now introduced are equivalent although in present form the latter is valid only for perfectly learned intact NNs (see details in ref. 8). Since for each $x_{in}$ exists such a vector $x(d)$ that $x_{in} = x(d)$, inequality $Q > \theta$ can be written as a function of $d = m/N$ and as a result

$$Q(d) = \sum_{i=1}^{N} x_0^i x_i(d) = \sum_{i=1}^{N-m} (x_0^i)^2 + \sum_{i=1}^{m} x_0^i x_r^i = N - m + (m - 2k) = N - 2k > \theta \tag{6}$$

where the dimension of all vectors $x$, the number of noise components of $x(d)$, the number of corresponding bits of $x(d)$ and $x_0$ which always coincide, and the number of corresponding bits of particular $x_r$ and $x_0$ which currently differ are $N$, $m$, $N - m$, and $k$, respectively; $\theta$ is threshold value of $Q$ or model neuron's triggering threshold.

It is easy to obtain directly that $Q = N - 2D$ and $D = (N - Q)/2$ where $D$ is a Hamming distance between $x_0$ and $x(d)$ (Hamming distance is a number of corresponding bits of $x_0$ and $x(d)$ which are different, $0 \leq D \leq N$). Since between $D$ and $Q$ there is an univocal correspondence, along with inequality $Q > \theta$ the inequality $D < (N - \theta)/2$ is also valid (cf. Inequality 6 where $k = D$). Moreover, $Q(d)$ can merely be interpreted as an expression for computation of Hamming distance $D$. That means that the above convolutional (Hamming) decoding algorithm or Hamming classifier directly discriminates the patterns $x_{in} = x(d)$ which are more close to $x_0$ than a given Hamming distance between them [8]. Hence, for data coding described in Section 2, NN, convolutional, and Hamming distance algorithms mentioned are equivalent. As Hamming classifier/recognition/decoding algorithm is the best (optimal) in the sense of statistical pattern recognition quality (that is no other algorithm cannot outperform it) [11], above NN and convolutional algorithms are optimal (the best) in that sense too.

## 4. Conditional Recall/Recognition Probabilities and ROCs

The basic idea of NNAMM is to build a NN memory model from simple objects defined within coding/decoding approach (optimal binary signal detection theory) introduced. For this purpose in Sections 2 and 3 it is simple enough instead of coding and decoding to speak about encoding and retrieval, respectively [8]. In this way NNAMM was formulated and fundamental recall/recognition properties of its assembly memory unit, containing corresponding Hopfield NN as its central element, were found optimally by multiple computations [7,8]. But convolutional (Hamming) version of NNAMM gives also a chance to obtain optimal (the best) formulae for this aim analytically.

Below we derive a formula for the probability $P(m,N,\theta)$ of correct recall/recognition of memory trace $x_0$ stored in perfectly learned intact NN with the model neurons' triggering threshold $\theta$ under condition that data patterns $x(d)$ initiating many-step memory trace retrieval [8] are actually $x_0$ with damage degree $d = m/N$ (earlier the same probability was calculated by multiple computations, examples for $\theta = 0$ see in ref. 7,8). Now we need to find the number $T(m,N,\theta)$ of vectors $x(d)$ for which Inequality 6 is valid and the total number of all possible different vectors $x(d)$. Since $x(d)$ contains $m$ randomly combined noise components with randomly chosen magnitudes –1 or 1 (the probability of their choice is ½), the latter equals $2^m C^N_m$. To find $T(m,N,\theta)$ we use the fact that for each set of $k$, $m$, and $N$ the number of vectors $x(d)$ satisfying Inequality 6 is $C^N_m C^m_k$ where $C^N_m$ is the number of ways arranging $m$ noise components in $N$ components of $x(d)$ and $C^m_k$ is the same for $k$ components which have the sign opposite to the sign of corresponding components of $x_0$ in $m$ noise components of $x(d)$. Consequently, $T(m,N,\theta) = C^N_m \sum C^m_k$ where the summation is made over $k = 0,1,..,kmax$ ($k$ is Hamming distance between particular $x(d)$ and $x_0$). The probability $P(m,N,\theta)$ is computed by dividing $T(m,N,\theta)$ by $2^m C^N_m$, i.e.

$$P(m,N,\theta) = \sum_{k=0}^{k\max} C_k^m / 2^m \qquad (7)$$

where if $kmax \leq kmax_0$ then $kmax = m$ else $kmax = kmax_0$ and

$$k\max_0 = \begin{cases} (N-\theta-1)/2, & \text{if } N \text{ is odd} \\ (N-\theta)/2 - 1, & \text{if } N \text{ is even} \end{cases} \qquad (8)$$

is defined by Inequality 6 and the signum function specified by Equation 5. Since $0 \leq kmax \leq m \leq N$, if $N$ is odd then $-(N+1) \leq \theta \leq N-1$ and if $N$ is even then $-(N+2) \leq \theta \leq N-2$.

Let us consider two important special cases, $P(m,N,\theta) = 1$ and $m = N$, $\theta = 0$:

- Since $\sum C^{kmax}_k = 2^{kmax}$ ($k = 0,1,..,kmax$), from Equation 7 follows that for any $N$ $P(m,N,\theta) = 1$ while $m \leq kmax_0$.
- Since $\sum C^N_k = 2^N$ ($k = 0,1,..,N$), if $N$ is odd then $P(N,N,0) = (2^N/2)/2^N = ½$ ($m = N$ and $\theta = 0$). Since $C^m_k = C^m_{m-k}$, if $N$ is even then the sum $S = \sum C^N_k$ ($k = 0,1,..,N/2 - 1$) is defined by equation $2S + C^N_{N/2} = 2^N$ ($C^N_{N/2}$ is the number of events $Q = 0$). Thus, in this case $P(N,N,0) = ½ - \Delta P(N)$, $\Delta P(N) = C^N_{N/2}/2^{N+1} \sim 0.4/\sqrt{N}$ (here for large $N$ Stirling's formula was used). The facts that $\Delta P(N) < 0$ and the minus sign was assigned to 1s in Expression 8 are caused by the choice of signum function form. If in Equation 5 for $h_j = \theta$ the value $v_j = +1$ is assigned then $\Delta P(N) > 0$ and in Expression 8 the plus sign before 1s should be chosen.

For odd and even $N$ and for different choice of signum function, probabilities $P(m,N,\theta = 0)$ are shown in Figure 1 (as in ref. 7,8 to underline discrete character of NNAMM results, small values of $N$ are taken, for example).

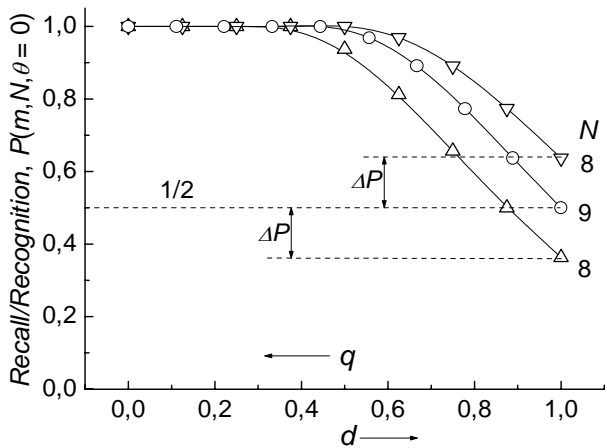

Figure 1. Conditional probability $P(m,N,\theta)$ of free recall ($d = 1$), cued recall ($0 < d < 1$), and recognition ($d = 0$) calculated according to Equations 7 and 8 for perfectly learned intact NNs with $\theta = 0$ and $N = 9$ (open circles) and $N = 8$ (triangles) vs. damage degree $d = m/N$ of memory trace $x_0$ or intensity of cue $q = 1 - m/N$. If $N = 9$ (i.e., $N$ is odd) then free recall (false alarm) probability equals ½; if $N = 8$ (i.e., $N$ is even) then free recall probability is $P(8,8,0) = ½ - \Delta P(8)$, $\Delta P(8) = C^8_4/2^9 = 70/512$. If $N = 9$ and $m \leq 4$, if $N = 8$ and $m \leq 3$ then $P(m,N,0) = 1$. If in Equation 5 for $h_j = \theta$ the value $v_j = 1$ is assigned then $P(8,8,0) = ½ + \Delta P(8)$ (upper curve).

Figure 2 shows two families of curves calculated according to Equations 7 and 8. In different form they represent the same probabilities $P(m,N,\theta)$ for perfectly learned intact NN memory unit with odd $N$ and all possible values of $d = m/N$ and $\theta$ (if $N$ is even then the curves can be splitted by the choice of signum function form).

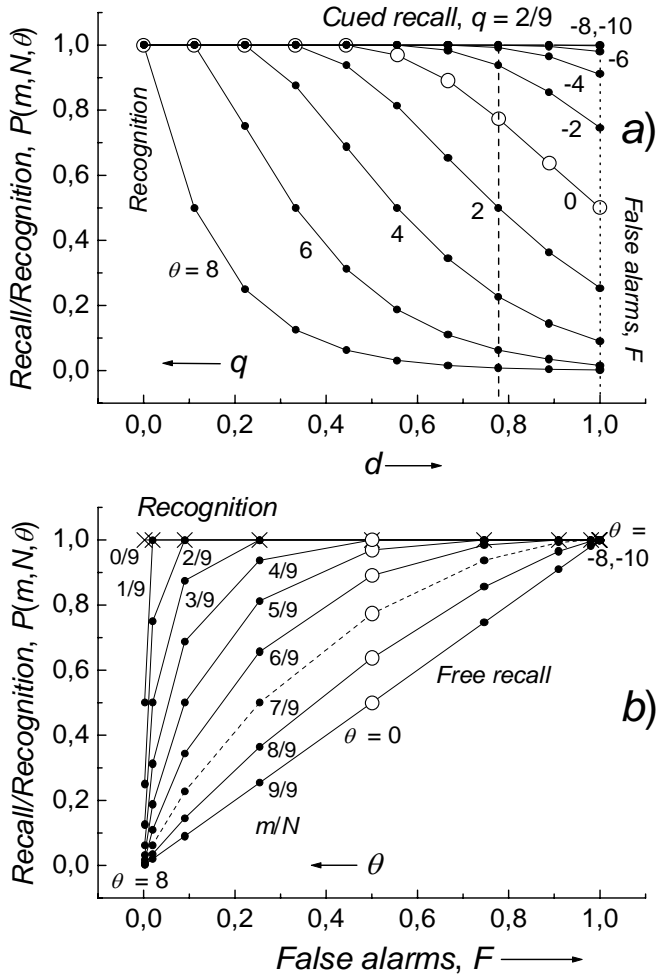

**Figure 2.** Data preparation for ROC plot (**a**) and ROCs (**b**) for the perfectly learned NN memory unit with $N = 9$. **a)** Probabilities $P(m,N,\theta)$ vs. $d = m/N$, $q = 1 - m/N$, and $\theta$. Open circles denote probabilities for $\theta = 0$, $P(m,9,0)$ (i.e., here and in all other Figures open-circle points are the same); dashed line connects cued recall probabilities for different $\theta$ and cue index $q = 2/9$, $P(7,9,\theta)$; free recall ($q = 0$) or false alarm probabilities, $F = P(9,9,\theta)$, are situated along dotted line; for all curves their left-most points are the same, $P(0,9,\theta) = 1$ (that is recognition probability). The number of curves is $N + 1 = 10$. Since $0 < F \le 1$, the value $F = 0$ is impossible. For each $\theta$ right-most point of each curve represents appropriate value of false alarm $F$ needed to plot ROCs. **b)** Probabilities $P(m,N,\theta)$ vs. $F$, $\theta$, and $m/N$. The values of $F$ used for the ROC plot lie in the panel a) along the dotted line. For each ROC curve the value of $m/N$ ($q$ or $d$) is the same, the number of ROC points is $N + 1 = 10$. The more the value of cue, $q$, the more the curvature of respective ROC and the more the value of probability $P(m,9,\theta = 8)$, ROC's left-most point. Linear ROCs correspond to free recall ($q = 0$, $d = 1$) and recognition ($q = 1$, $d = 0$). Crosses denote recognition probabilities, $P(0,9,\theta)$.

## 5. Unconditional Recall/Recognition Probabilities, mROCs, and Overall Probabilities

In Section 4 conditional recall/recognition probabilities were discussed. But it is *a priori* unknown whether initial pattern $x(d)$ is a sample of noise (hypothesis $H_0$) or memory trace $x_0$ damaged by noise (hypothesis $H_1$). To obtain unconditional (*a posteriori*) probabilities of false and correct recall/recognition of the trace $x_0$ stored in NN memory unit, we used famous Bayes' formula and have as a result:

$$p_{FR}(m/N,F) = 1/(1+\kappa \frac{P(m/N,F)}{F}), \quad p_{CR}(m/N,F) = 1/(1+\kappa^{-1}\frac{F}{P(m/N,F)}) \tag{9}$$

where $p_{FR}(m/N,F)$ and $p_{CR}(m/N,F)$ reflect unconditional false recall/recognition (FR) and correct recall/recognition (CR) probabilities; $p_{FR} + p_{CR} = 1$; $\kappa = P(H_1)/P(H_0)$; $P(H_0)$ and $P(H_1)$ are prior probabilities of hypotheses $H_0$ and $H_1$, respectively. Since $P(H_0)$ and $P(H_1)$ are usually unknown, in most cases $\kappa = 1$ is postulated. Here, we pay also attention to the fact of changing designations. As there is an univocal correspondence between $F$ and $\theta$ (see Figure 2a) in Equation 9 and below, instead of $\theta$, we write $F$; as all probabilities depend on $m$ and $N$ as on $m/N$ (see Figures 1 and 2), we write these two parameters in the form of their ratio; $P(m/N,F) = P(m,N,\theta)$.

Our data coding approach introduced in ref. 7 allows to find $\kappa$ in explicit form directly. Indeed, by definition, a pattern $x(d)$ contains a fraction $d = m/N$ of noise components and a fraction $q = 1 - m/N$ of undamaged components of $x_0$ (see Section 2). Hence, $d$ and $q$ may be interpreted as the probabilities $P(H_0)$ and $P(H_1)$, respectively. That means that in Equation 9, within our NNAMM (or data coding/decoding) approach, $\kappa$ is given by

$$\kappa = P(H_1)/P(H_0) = q/d = (N-m)/m. \qquad (10)$$

If $m = 0$ then, according to Equation 10, $\kappa$ does not exist and in this special case we posit that $p_{FR} = 0$ (at the same time $p_{CR} = 1$); if $m = N$ then $\kappa^{-1}$ does not exist and in this special case we posit that $p_{CR} = 0$ (at the same time $p_{FR} = 1$). Both propositions are in full concordance with the fact that the former is the case of undamaged memory trace $x_0$ and the latter is a case of pure noise. Taking into account that $0 < F \leq P(m/N,F) \leq 1$ (see Figure 2), that Equation 10 and propositions $p_{FR} = 1$ (if $m = 0$), $p_{CR} = 0$ (if $N = m$) are valid, we have: $0 \leq p_{FR} \leq 1$, $0 \leq p_{CR} \leq 1$ (instead of $0 < p_{FR} \leq \frac{1}{2}$, $\frac{1}{2} \leq p_{CR} < 1$ if it is as usual supposed that $\kappa = 1$ and $0 < F \leq P(m/N,F) < 1$).

Equations 9,10 provide unconditional probability $p_{CR}(m/N,F)$ as a function of false alarm $F$ and for this reason for the fixed $m/N$ we refer to particular $p_{CR}(m/N,F)$ as modified ROC or mROC. Figure 3 illustrates this claim.

Let us define

$$P_{FR}(m/N) = \sum p_{FR}(m/N,F)/(N+1), \quad P_{CR}(m/N) = \sum p_{CR}(m/N,F)/(N+1) \qquad (11)$$

where $P_{FR}(m/N)$ and $P_{CR}(m/N)$ provide overall, not depending on $F$, unconditional FR and CR probabilities of recall/recognition of the memory trace $x_0$ stored in perfectly learned NN; summations are made over all $0 < F \leq 1$; $p_{FR}(m/N,F)$, $p_{CR}(m/N,F)$ are calculated according to Equation 9; as it was expected, $P_{FR} + P_{CR} = 1$.

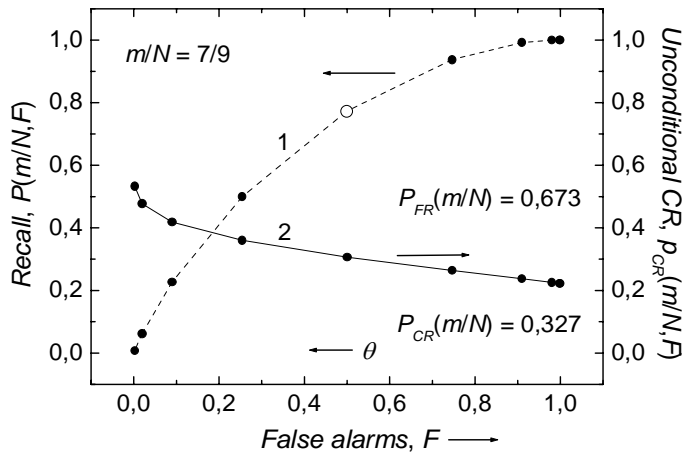

Figure 3. ROC curve (curve 1, left-hand scale) and mROC curve (curve 2, right-hand scale) for $d = m/N = 7/9$, $q = 2/9$. ROCs along the dashed line are as in Figure 2. The mROC curve is a plot of unconditional correct recall probability $p_{CR}(m/N,F)$ vs. false alarm $F$; mROC points according to Equations 9,10 were calculated; the special case $p_{CR}(m/N,F) = 1$ is not shown and not considered. Average values of $p_{CR}(m/N,F)$ and $p_{FR}(m/N,F)$ reflect overall probabilities $P_{FR}(m/N)$ and $P_{CR}(m/N)$, respectively, they are estimated by Equations 11.

## 6. Comparison with Experiments

In Figure 4 NNAMM numerical predictions (calculated ROC curves) are compared with ROCs observed in item recognition or similar tests. In different panels typical examples of empirical many-point and two-point ROCs are examined, estimated empirical data were taken from ref. 3. As one can see, even for illustrative model example $N = 9$ where only cue index $q$ was as a fit parameter (the change of $N$ does not change the form of ROCs), a good quantitative agreement between theory and experiment is achieved. Thus, the comparison of empirical and model ROCs may be viewed as a method for estimation of specific value of the intensity of cue available in the process of the recall or recognition for specific memory system under specific conditions of specific experiment.

As Figure 4 demonstrates, there is no problem of reproducing available empirical ROCs within NNAMM both qualitatively and quantitatively and comparison of calculated and empirical ROCs may be successfully used for the value of intensity of cue, $q$, estimation. Since for empirical many-point-confidence-scale ROCs the value of cue changes along the curves (see Figure 4a), both the model's predictions and the details of experimental protocols demand scrutiny.

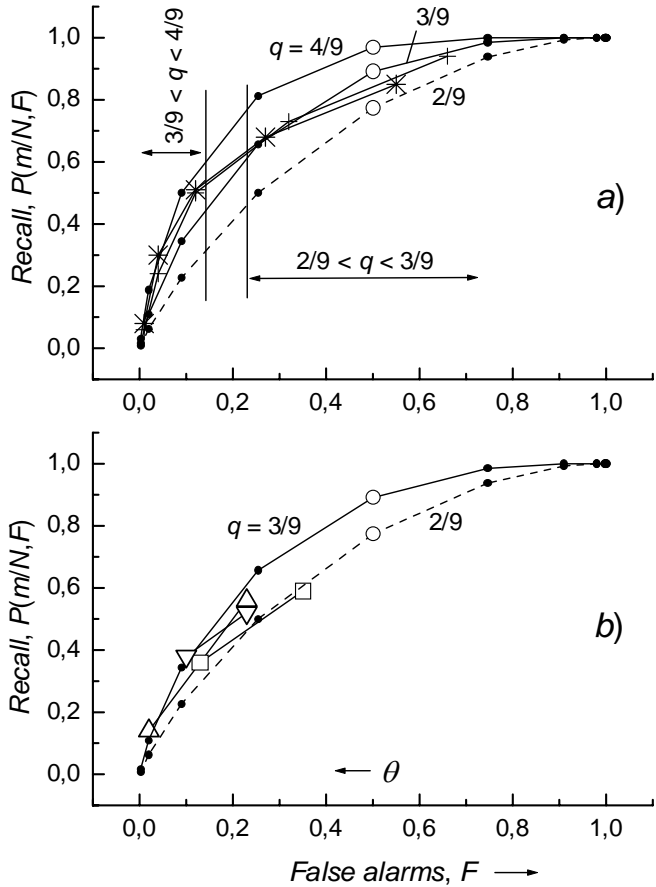

Figure 4. Theoretical ($N = 9$) and empirical ROCs. For each calculated ROC respective value of $q$ is shown. Here and in Figures 2 and 3 dashed-line curve is the same. a) Comparison of theoretical and empirical ROCs derived using 5-point-confidence-scale experiments. Original results are from ref. 12 and 13, the first 3 and the last 2 points of empirical ROCs are consistent with the assumption that $3/9 < q < 4/9$ and $2/9 < q < 3/9$, respectively. b) The same for empirical ROCs derived using 2-point-confidence-scale experiments. Original results are from ref. 14 and 15, they are consistent with $2/9 < q < 3/9$.

In many experiments (e.g., associative recognition test, remember/know or process-dissociation procedures) subjects are required to recall both an item itself and other information related to it [3]. That means that in such experiments those memory events could be selected and investigated where subjects are able a target item to retrieve and "to assess" its *a posteriori* probability taking into account *a priori* probabilities of prior hypotheses on the structure of information initiated retrieval (i.e., taking into account $P(H_1)$, the probability of the fact that vector $x(d)$ reflects damaged target item, and $P(H_0)$, the probability of the fact that $x(d)$ is a lure item). Hence, empirical results obtained using such an experimental paradigm could provide unconditional (*a posteriori*) recall/recognition probabilities $p_{CR}(m/N,F)$ introduced in Section 5. This assumption is examined in Figure 5.

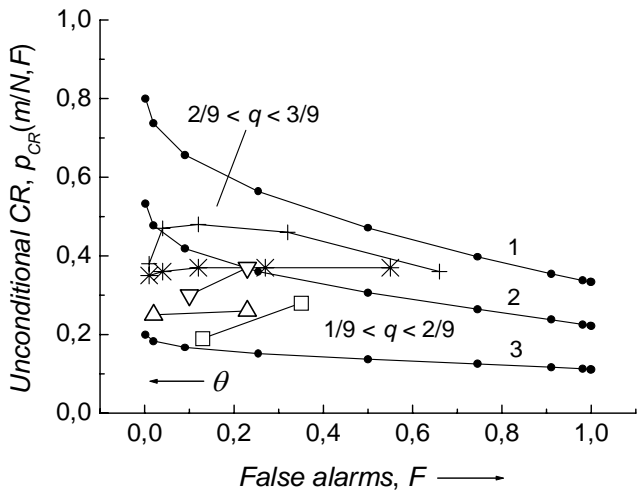

Figure 5. Theoretical ($N = 9$) and empirical mROCs. Curves 1, 2, and 3 reflect $p_{CR}(m/N,F)$ calculated according to Equations 9,10 with cue indices $q = 1 - m/N = 3/9$, $2/9$, and $1/9$, respectively. Empirical mROC curve in the same signs as ROC curve in Figure 4 was taken from the same reference [3,12-15].

As Figure 5 demonstrates, theoretical mROCs provide good quantitative description of observed mROCs [3,12-15] and their comparison may also be viewed as a method for estimation of specific values of the intensity of cue for specific memory experiments. For example, empirical 2-point mROCs [14,15] and 5-point mROCs [12,13] are consistent with the assumption that $1/9 < q < 2/9$ and $1/9 < q < 2/9$ to $2/9 < q < 3/9$, respectively (in the latter case $q$ changes along the curves).

Comparison between the values of $q$ estimated using ROCs and mROCs shows that they are similar but not always coincide. Indeed, an analysis of ROCs and mROCs observed in experiments [14,15] gives inconsistent results ($2/9 < q < 3/9$ and $1/9 < q < 2/9$, respectively) while the analysis of experiments [12,13] gives consistent results if only 2 right-most ROC and mROC points are considered (2/9 < $q$ < 3/9) and inconsistent results if some left-most ROC and mROC points are taken into account (3/9 < $q$ < 4/9 and 1/9 < $q$ < 2/9, respectively). To explain these features, additional analysis of the model's predictions and experimental details is needed.

## 7. Discussion

The properties of empirical ROC curves have been used as one of four basic arguments in favour of DPMs of recognition memory. For example, as Figures 4 and 5 demonstrate empirical ROCs derived in item and associative recognition tests are essentially different [3]. ROCs related to item recognition tests are curvilinear with changing shape across measurement conditions; they can be approximated by a two-factor formula related to traditional signal detection theory and containing recollection and familiarity as stochastically independent fit parameters. For this reason, it is claimed that "at least two separate memory components are needed to account for recognition performance" [3, p.442]. This idea was realized as a two-factor parameterization of empirical ROCs: $P_i = R + (1 - R)\Phi(d'/2 - c_i) + F_i - \Phi(d'/2 - c_i)$ where $P_i$, $F_i$, $R$, $d'$, $c_i$, and $\Phi$ reflect correct recall probability (a counterpart to probability $P(m/N,F)$ defined by Equation 7), false alarm, recollection, familiarity, response criterion, and item distribution (Gaussian), respectively. The fitting of this equation to observed ROC curves provides estimations of recollection ($R$) and familiarity ($d'$). Since ROCs observed in item recognition tests (Figure 4) are well fitted by this formula and ROCs observed in associative recognition tests (Figure 5) are not, it is suggested that the former can be described by a signal detection theory while the latter can not [3].

Our NNAMM is based on our optimal binary signal detection theory (Sections 2-5, ref. 7-9) and for intact perfectly learned memory unit it is actually a one-factor computer model; this factor (intensity of cue or cue index, $q$) is the amount of undamaged information about the memory trace $x_0$ containing in vectors $x(d)$ which initiate many-step memory retrieval [8]. It is essential that such an one-factor approach on a common ground successfully describes different types of memory including free recall ($q = 0$), cued recall ($0 < q < 1$), and recognition ($q = 1$) [7,8] and for this reason there is no need to introduce any new type of memory, like recollection or familiarity, for example ("recollection" and "familiarity" of DPMs are loosely equivalent to recall and recognition of NNAMM, respectively). By definition, all acts of the particular item's recall and recognition are different in time processes and, consequently, they are stochastically independent and do not run in parallel. According to NNAMM, recognition ("familiarity" of DPMs) is an one-step process of testing selected assembly memory unit (AMU) *without* using the cues stored in other related AMUs [8]. In general, such a process can correspond to an item recognition test of so called *semantic* memory. Recall ("recollection" of DPMs) is a many-step process of testing selected AMU *with* using the cues stored in other (one or more) AMUs [8]. In general, such a process can correspond to an associative recognition test of so called *episodic* memory (for relations between semantic and episodic memories see ref. 16, for example). As one can see from Section 6, ROCs observed in item recognition tests and mROCs observed in associative recognition tests are successfully described within our NNAMM based on our optimal binary signal detection theory.

Since all basic properties of empirical ROCs (and mROCs) have been qualitatively and even quantitatively reproduced within one-factor NNAMM, ROCs might be excluded from the list of findings underpinning DPMs of recognition memory. On the ground of our present and previous [7,8,17,18] results it is natural to anticipate other items of this list (different speeds of response for recollection and familiarity, their different electrophysiological correlates, and different extents of their disruption by certain brain injuries) are also consistent with NNAMM.

## 8. Conclusion

For the first time a method for theoretical description of empirical ROC curves has been proposed within a computer memory model. For this purpose a convolutional (Hamming) version of our NNAMM based on our optimal binary signal detection theory was used. Analytical formulae for optimal (the best) calculation of conditional and unconditional probabilities of false/correct recall/recognition of memory trace stored in intact perfectly learned NN memory unit have been found. In particular, a method of taking into account explicitly *a priori* probabilities of alternative hypotheses on the structure of information, i.e. vectors $x(d)$, initiated memory retrieval and a method for estimation of overall recall/recognition probabilities are proposed. Using the derived optimal analytical formulae, empirical ROCs obtained in item recognition tests and empirical mROCs obtained in associative recognition tests were described and the values of intensity of cue, $q$, for some specific experiments were quantitatively estimated; thus, the comparison of theoretical and empirical ROCs is a method proposed here to estimate cue indices for specific experiments. It has been shown that ROCs might be excluded from the list of empirical findings underpinning popular DPMs of recognition memory.

I am grateful to HINARI (Health Internetwork Access to Research Initiative) for free on-line access to recent full-text journals, participants of the KDS-2003 Conference, Varna, Bulgaria, June 16-26, 2003 for helpful discussion, and my family and my friends for their help and support.

## Bibliography

[1] D.Green & J.Swets. Signal Detection Theory and Psychophysics. New York, Wiley, 1966.

[2] W. Metzner, C. Koch, R.Wessel, & F.Gabbiani. Feature Extraction by Burst-Like Spike Pattern in Multiple Sensory Maps. Journal of Neuroscience, 1998, **18**(6), 2283-2300.

[3] A.P.Yonelinas. The Nature of Recollection and Familiarity: A Review of 30 Years of Research. Journal of Memory and Language, 2002, **46**, 441-517.

[4] B.B.Murdock. Classical Learning Theory and Neural Networks. The Handbook of Brain Theory and Neural Networks. Cambridge, MIT Press, 1995, 189-192.

[5] J.R.Manns, R.O.Hopkins, J.M.Reed, & E.G.Kitchener, L.R.Squire. Recognition Memory and the Human Hippocampus. Neuron, 2003, **37**(1), 171-180.

[6] K.A.Norman & R.C.O'Reilly. Modeling Hippocampal and Neocortical Contributions to Recognition Memory: A Complementary Learning Systems Approach. Psychological Review, 2003, **110**(4).

[7] P.M.Gopych. Determination of Memory Performance. JINR Rapid Communications, 1999, **4**[96]-99, 61-68 (in Russian).

[8] P.M.Gopych. A Neural Network Assembly Memory Model with Maximum-Likelihood Recall and Recognition Properties. Vth International Congress on Mathematical Modeling, Dubna, Russia, September 30 – October 6, 2002, Book of Abstracts, vol.2, p.91. See also http://arXiv.org/abs/cs.AI/0303017.

[9] P.M.Gopych. Identification of Peaks in Line Spectra Using the Algorithm Imitating the Neural Network Operation. Instruments & Experimental Techniques, 1998, **41**(3), 341-346.

[10] J.J.Hopfield & D.W.Tank. Computing with Neural Circuits: a Model. Science, 1986, **233**, 625-633.

[11] R.Hamming. Coding and Information Theory. Englewood Cliffs, NJ, Prentice Hall, 1986.

[12] A.P.Yonelinas. Receiver-Operating Characteristics in Recognition Memory: Evidence for a Dual-Process Model. Journal of Experimental Psychology: Learning, Memory, and Cognition, 1994, **20**(6), 1341-1354.

[13] A.P.Yonelinas. Consciousness, Control, and Confidence: The 3 Cs of Recognition Memory. Journal of Experimental Psychology: General, 2001, **130**(3), 361-379.

[14] F. Strack & J.Foerster. Reporting Recollective Experiences: Direct Access to Memory Systems? Psychological Science, 1995, **6**(6), 352-358.

[15] E.Hirshman & A.Henzler. The Role of Decision Processes in Conscious Recollection. Psychological Science, 1998, **9**(1), 61-65.

[16] E.Tulving, H.J.Markovitsch. Episodic and Declarative Memory: Role of the Hyppocampus. Hyppocampus, 1998, **8**, 198-204.

[17] P.M.Gopych. Three-Stage Quantitative Neural Network Model of the Tip-Of-the-Tongue Phenomenon. Proceedings of the IXth International Conference KDS-2001, St-Petersburg, Russia, June 19-22, 2001, p.158-165 (in Russian). See also http://arXiv.org/abs/cs.CL/0103002, http://arXiv.org/abs/cs.CL/0107012, http://arXiv.org/abs/cs.CL/0204008.

[18] P.M.Gopych. Computer Modeling of Feelings and Emotions: A Quantitative Neural Network Model of the Feeling-Of-Knowing. Kharkiv University Bulletin, Series Psychology, 2002, no.550(1), 54-58 (in Russian). See also http://arXiv.org/abs/cs.AI/0206008.

## Author information

**Petro Mykhaylovych Gopych** – V.N.Karazin Kharkiv National University; Svoboda Sq., 4, Kharkiv, 61077, Ukraine; e-mail: pmg@kharkov.com.